\def\year{2021}\relax
\documentclass[letterpaper]{article} % DO NOT CHANGE THIS

\usepackage{aaai21}  % DO NOT CHANGE THIS
\usepackage{times}  % DO NOT CHANGE THIS
\usepackage{helvet} % DO NOT CHANGE THIS
\usepackage{courier}  % DO NOT CHANGE THIS
\usepackage[hyphens]{url}  % DO NOT CHANGE THIS
\usepackage{graphicx} % DO NOT CHANGE THIS
\usepackage{natbib}  % DO NOT CHANGE THIS AND DO NOT ADD ANY OPTIONS TO IT
\usepackage{caption} % DO NOT CHANGE THIS AND DO NOT ADD ANY OPTIONS TO IT
\usepackage[ruled,vlined,linesnumbered,noend]{algorithm2e}
\usepackage{booktabs}
\usepackage{xcolor}
\graphicspath{{figures/}}
\usepackage{amsmath}
\usepackage{float}

\SetCommentSty{mycommfont}
\usepackage[export]{adjustbox}
\usepackage{subcaption}
\usepackage{dblfloatfix}
\usepackage[normalem]{ulem}
\usepackage[all]{nowidow}

\title{Robust Blockchained Federated Learning \\ with Model Validation and Proof-of-Stake Inspired Consensus}
\author {
    Hang Chen,\textsuperscript{\rm 1} 
    Syed Ali Asif,\textsuperscript{\rm 1} 
    Jihong Park,\textsuperscript{\rm 2}
    Chien-Chung Shen,\textsuperscript{\rm 1}
    Mehdi Bennis\textsuperscript{\rm 3}\\
}
\affiliations {
    \textsuperscript{\rm 1} Department of Computer and Information Sciences, University of Delaware, USA \\
    \textsuperscript{\rm 2} School of Information Technology, Deakin University, Australia\\
    \textsuperscript{\rm 3} Centre for Wireless Communications, University of Oulu, Finland\\
    chenhang,asifrabi,cshen@udel.edu, jihong.park@deakin.edu.au, mehdi.bennis@oulu.fi
}
\begin{document}

\maketitle

\begin{abstract}
Federated learning (FL) is a promising distributed learning solution that only exchanges model parameters without revealing raw data. However, the centralized architecture of FL is vulnerable to the single point of failure. In addition, FL does not examine the legitimacy of local models, so even a small fraction of malicious devices can disrupt global training. To resolve these robustness issues of FL, in this paper, we propose a blockchain-based decentralized FL framework, termed VBFL, by exploiting two mechanisms in a blockchained architecture. First, we introduced a novel decentralized validation mechanism such that the legitimacy of local model updates is examined by individual validators. Second, we designed a dedicated proof-of-stake consensus mechanism where stake is more frequently rewarded to honest devices, which protects the legitimate local model updates by increasing their chances of dictating the blocks appended to the blockchain. Together, these solutions promote more federation within legitimate devices, enabling robust FL. Our emulation results of the MNIST classification corroborate that with $15$\% of malicious devices, VBFL achieves $87$\% accuracy, which is $7.4$x higher than Vanilla FL.

\end{abstract}

\section{Introduction}
%COMMENT\textcolor{cyan}{[MB: we need to compare pros and cons of PoW and PoS in terms of communication latency and compute.]}\textcolor{blue}{[Hang: Added in the experimental results in supplementary material]}
% [PIEEE1]: Wireless Network Intelligence at the Edge
% [PIEEE2]: Communication-Efficient and Distributed Learning Over Wireless Networks: Principles and Applications
% [Network]: Distilling On-Device Intelligence at the Wireless Network Edge
% [Google]: Advances and Open Problems in Federated Learning
% [FedML] FedML: A Research Library and Benchmark for Federated Machine Learning
% [McMahan]: Communication-Efficient Learning of Deep Networks from Decentralized Data
% [BlockFL]: \cite{kim2019blockchained}
% [GADMM] GADMM: Fast and Communication Efficient Framework for Distributed Machine Learning
% [Jaggi]: Decentralized Stochastic Optimization and Gossip Algorithms with Compressed Communication
% [PoS]: Cryptocurrencies without proof of work
% [REFs]: Add relevant blockFL workers we're referring to

Fundamental requirements of machine learning rest on securing a large volume of high-quality data that accurately reflects the current environments or situations. In search of the sheer amount of fresh data, networked edge devices, such as phones, cameras, and Internet of Things (IoT), from which the freshest data are generated have recently attracted significant attention \cite{park2019wireless,park2019distilling,park2020communication}. However, collecting their raw data brings about non-negligible latency and security issues, as these user-generated data are dispersed across devices and often privacy sensitive, e.g., location history and health records. In this respect, federated learning (FL) is a convincing solution that periodically exchanges local model parameters, rather than instantaneously exchanging raw data \cite{mcmahan2017communication,kairouz2019advances,he2020fedml}. In essence, FL constructs a global model by aggregating local models at a central server. While effective under benign environments, this simple operational principle of FL leads to two robustness issues in reality as elaborated next. 

\begin{figure}
    \centering
    \includegraphics[width=0.9\columnwidth, bb=0 0 1371 734]{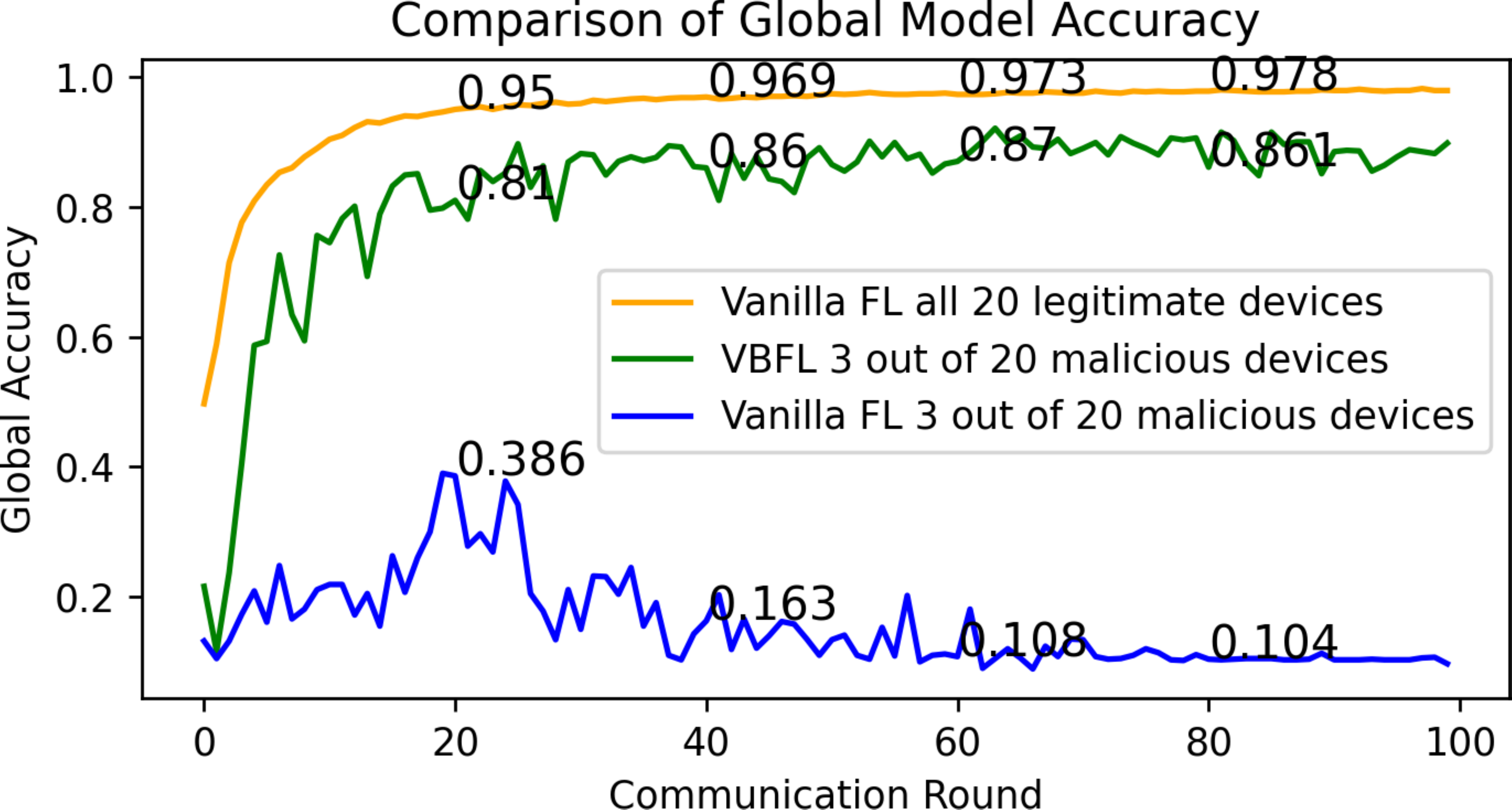}
    \caption{Global model accuracy comparison between Vanilla FL and VBFL, with and without malicious devices in the MNIST classification. 
    %COMMENT\textcolor{red}{[MB: this result seems too early in the paper.]}
    } 
    \label{fig:f1_vfl_VBFL_comparison}
\vspace*{-0.2in}
\end{figure}

First, FL is vulnerable to the single point of failure due to its central parameter server aggregating local models to construct a global model \cite{bonawitz2019towards,elgabli2020gadmm}.
%Exploiting decentralized consensus algorithms can mitigate this problem \cite{koloskova2019decentralized,lyu2020towards}, which comes at the cost of compromising communication efficiency. 
To address this issue, blockchain-based FL is one promising decentralized solution, yet with a large number of devices, its proof-of-work (PoW) consensus delays are significant \cite{kim2019blockchained}. Second, FL is not capable of filtering out malfunction or malicious attacks during the global model construction. As observed by our MNIST experiments depicted in Fig.~\ref{fig:f1_vfl_VBFL_comparison}, the global model accuracy of the most basic form of FL, termed Vanilla FL, can collapse with only $15$\% of malicious devices injecting Gaussian noise into their local models. A na\"ive solution is to compare the accuracy evaluated on each local model with others \cite{sheller2020federated}. However, such a method is not viable under a decentralized architecture due to the lack of models for comparison, because not all local models can be easily collected at a single location within the limited time.

\begin{figure}
\begin{subfigure}[b]{0.45\textwidth}
   \centering
   \includegraphics[width=0.9\linewidth]{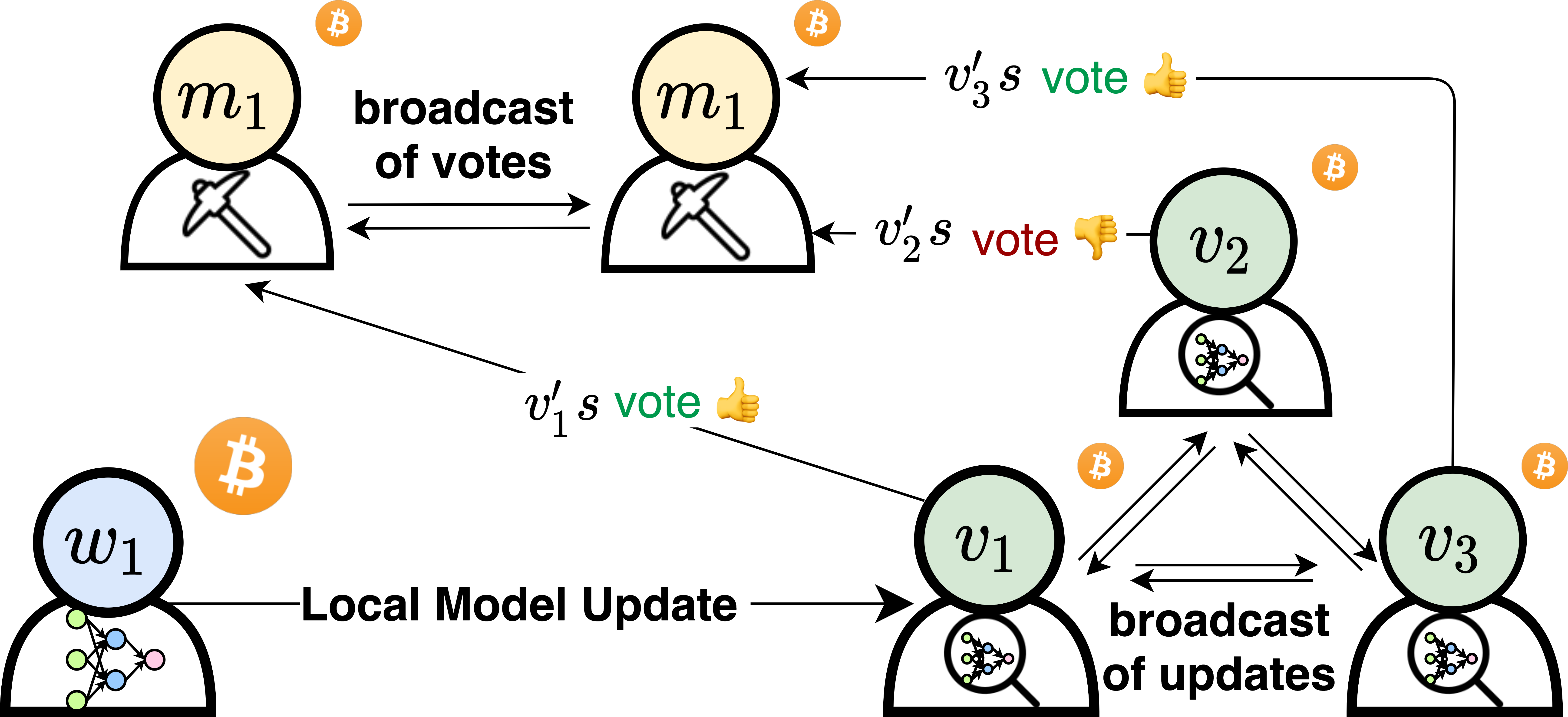}
   \caption{Voting and transmission of local model updates.}
\end{subfigure}
\par\bigskip
% \vspace*{0.5in}
\begin{subfigure}[b]{0.45\textwidth}
   \centering
   \includegraphics[width=0.9\linewidth]{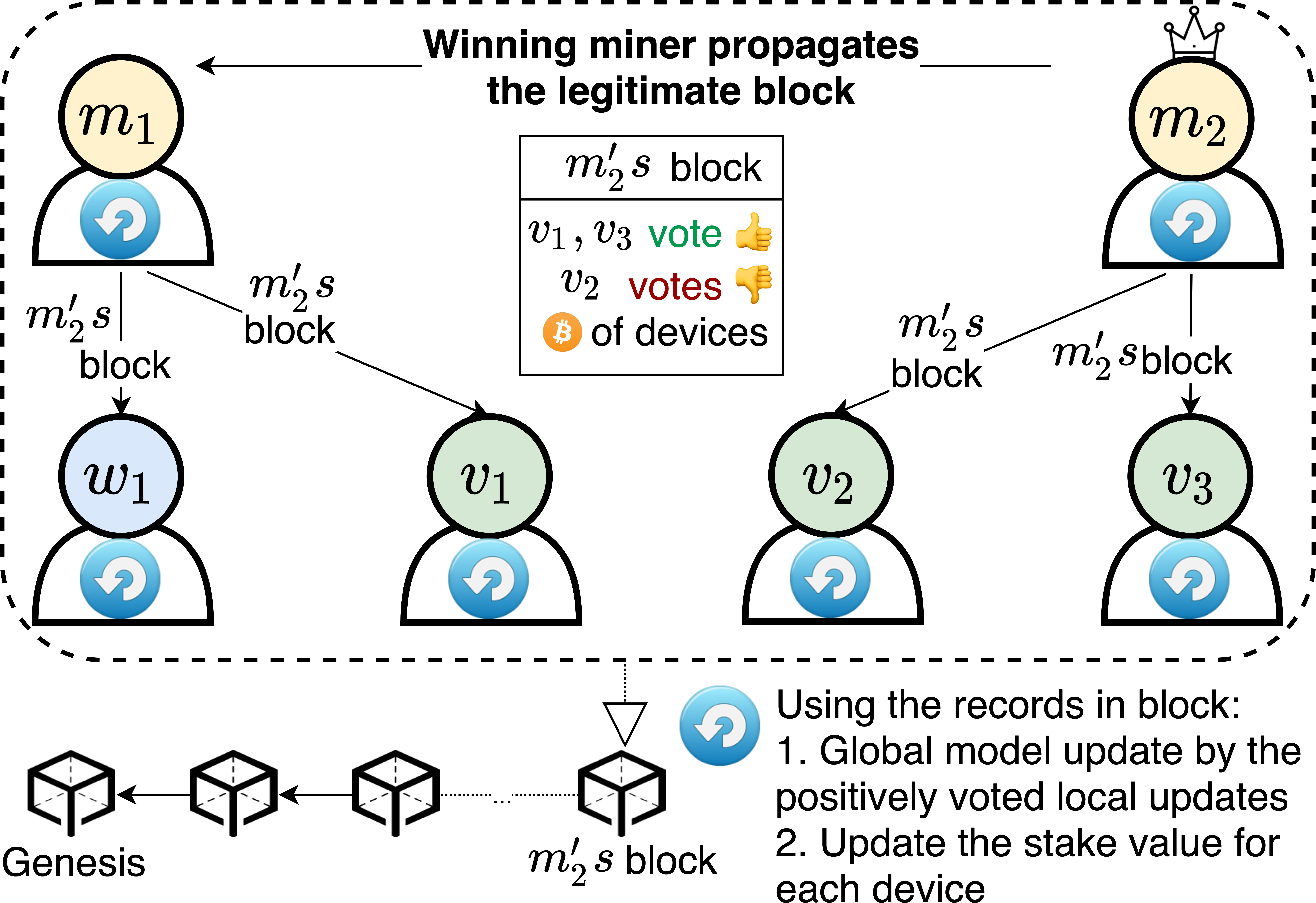}
   \caption{Voting results aggregation, block generation and block processing (i.e., global model and stake information update).}
\end{subfigure}
\caption{VBFL operations in one communication round.}
\label{fig:overall}
% \vspace*{-0.1in}
\end{figure}

To resolve the two aforementioned robustness issues of FL, 
%without compromising communication efficiency, 
we propose a novel \emph{blockchained FL framework}, coined VBFL, by introducing (1) \emph{a validation mechanism that examines and votes on local model updates in a decentralized manner} and (2) \emph{a proof-of-stake (PoS) inspired consensus mechanism that allows devices making the most learning contributions to create legitimate blocks recording local model updates with their corresponding voting results}. In VBFL, at every communication round, each device possessing its local dataset is randomly appointed as: (i) a machine learning \emph{worker} updating its local model, (ii) a model \emph{validator} examining and voting on the legitimacy of the received local model updates, or (iii) a blockchain \emph{miner}\footnote{In Ethereum \cite{buterin2014next} and other PoS-based blockchains, it is common that validators cast votes for generating a consensual block. In VBFL, to distinguish the role of voting for legitimate models and that of creating consensual blocks, validators and miners are used, respectively.} attempting to incorporate voting results with the corresponding local models to be stored in the next consensual block. Given this role-switching policy, VBFL aims to promote more federation within legitimate devices via the following complementary mechanisms. 
\begin{itemize}
    \item \textbf{\textcolor{black}{Decentralized} Model Validation}: Once a local model update is received by a validator, the validator broadcasts this update to other validators and runs training using its own local dataset for a single epoch to evaluate the accuracy of the received model. From our experimental observations, illegitimate models of malicious devices exhibit noticeably steep accuracy drops after such extra training due to their maliciously distorted models. By exploiting this phenomenon, a validator casts its vote on the legitimacy of each model. Based on the accumulated votes from multiple validators, potential malicious devices associated with illegitimate models are kicked out from the VBFL operations. \textcolor{black}{Moreover, our experiments show that a few compromised validators cannot disrupt the validation results as the validating operations are decentralized.}
    \item \textbf{PoS-inspired Consensus}: The consensus mechanism of VBFL is inspired by proof-of-stake (PoS) \cite{king2012ppcoin}, which allows the device serving the role of miner that has made the most learning contributions among the assigned miners in each communication round (i.e., the device that has been mostly cumulatively rewarded among the devices that are assigned to miners) to create the legitimate block recording local model updates with their corresponding voting results.
    % in which each global model is constructed by a winning miner whose winning probability is proportional to the miner's accumulated reward (i.e., stake). 
    % A reward is associated with each device, and is provided for its every role of (i), (ii), or (iii), which is recorded in the blockchain and recognized by the devices. Owing to the validation mechanism, malicious devices are less likely to frequently receive the reward for (i), which in turn decreases the reward for (iii).
    Rewards are granted to each device, and become a part of the stake of the device, based on the role of (i), (ii), or (iii) it plays in each round. The stake of each device is recorded in the blockchain and recognized by the devices in the network. Owing to the validation mechanism, malicious devices are less likely to frequently receive the rewards for playing workers, which in turn decreases their accumulated rewards (i.e., stake), thus mitigating their chance of being selected as the winning-miner. \textcolor{black}{Additionally, our experiments suggest that the proposed PoS-inspired consensus is much more communication-efficient compared to the PoW-based approach due to the shorter block generation time, while the frequency of forking events is still comparable according to our emulation results.}
\end{itemize}
A high-level overview of the described VBFL operations involving 1 worker, 3 validators and 2 miners is depicted in Fig.~\ref{fig:overall}. Consequently, even with $15$\% of malicious devices, Fig.~\ref{fig:f1_vfl_VBFL_comparison} shows that VBFL achieves $87$\% accuracy, which is $7.4$x higher than Vanilla FL without the proposed validation and PoS enhancements. Note that VBFL is the first PoS-based blockchained FL system practically implemented using Python, in contrast to existing blockchained FL works \cite{kang2020reliable,li2020blockchain,awan2019poster,qu2019proof,bonawitz2019towards,ma2020federated} that commonly rely on mathematical modeling and/or algorithmic simplifications (e.g., synchronous mining operations \cite{kim2019blockchained}). Our VBFL %implementation, emulation 
prototype is open-source and available at: \texttt{https://github.com/hanglearning/VBFL}.

\section{Related Work}
%COMMENT\textcolor{cyan}{[MB: add the P. Kairouz et al paper "advances in FL".]}\textcolor{blue}{added}
FL addressed research issues of data privacy and data heterogeneity for collaborative machine learning (ML) \cite{kairouz2019advances,he2020fedml}. To handle the issue of single-point-of-failure in FL, blockchain decentralizes the aggregation of local updates \cite{ramanan2019baffle,awan2019poster,ma2020federated}. Moreover, blockchain-based FL facilitates traceability, auditability, and tamper-proof to ensure reliable and trustworthy learning processes \cite{bonawitz2019towards,yuan2020federated,lyu2020towards}. Blockchain-based Smart Contracts (SC) were used \cite{ ramanan2019baffle} to eliminate limitations of aggregators in the FL environment using a large Deep Neural Network (DNN) model. Furthermore, using blockchain, one study \cite{lyu2020towards} described a local credibility mutual evaluation mechanism for fairness and a multi-layer encryption scheme for accuracy and privacy for DL models. However, the works did not focus on the consensus protocols regarding poisoning attacks, which left room for performance degradation and vulnerability if there exist malicious and/or compromised nodes in the system uploading noisy local updates for global model construction.

An approach was devised \cite{kang2020reliable} to guarantee reliability by utilizing the reputation of the contributing nodes to select trusted worker nodes for enhancing the performance of model training. The inspiration came from reputation usage in solving the data quality problems in crowdsensing and the multi-weighted subjective logic model utilizing the degree of belief, distrust, and uncertainty to compute reputation. However, the storage of reputation into a separate consortium blockchain results in overhead \cite{kang2020reliable}. A committee consensus for a decentralized blockchain-based DL framework \cite{li2020blockchain} was proposed to deal with the malicious worker updates. In the approach, the authors constructed a committee overseeing and testing the worker nodes' local model updates by validating with their own data and calculating a validation score. The authors followed the Byzantine Fault Tolerance (BFT) protocol for safeguarding the operations of the method \cite{li2020blockchain}.

The theoretical studies by the authors \cite{kim2019blockchained} planned a proof-of-work (PoW) based on-device FL architecture for DL applications by analyzing end-to-end latency and deriving optimal block generation rate for better scalable performance. An energy recycling Proof of Federated Learning (PoFL) approach with reverse game-based data trading and a privacy-preserving model verification method overcame latency drawback of PoW and data privacy leakage during model training and verification in FL but failed to address the poisoning attacks \cite{qu2019proof}. Other studies such as \cite{fung2018mitigating} targeted specific poisoning attacks (Sybil Attacks) on FL and proposed mitigation strategy based on dissecting the diversity of the local model updates. Specifically, too many similar updates from a group of nodes are labeled as malicious and treated accordingly. Additionally, in \cite{ liu2020secure} a central aggregator recognized malicious and unreliable participants by automatically executing smart contracts to defend against poisoning attacks, and various local differential privacy techniques were used to prevent membership inference attacks on the system \cite{ liu2020secure}. Nevertheless, complications from having a central aggregator such as bias towards selective worker nodes \cite{ramanan2019baffle} could still corrupt the FL process, which was addressed and eliminated by the role-switching policy in our approach.  

In a comparative analysis of consensus algorithms in blockchain, the authors discussed PoW, PoS, and BFT algorithms \cite{bach2018comparative}. PoW \cite{kim2019blockchained} based approach incurs more latency and communication overhead. BFT based approaches \cite{li2020blockchain} work in the permissioned blockchains and affect the anonymity of the participant nodes. Moreover, the reported results achieved 90\% accuracy when there are 50\% malicious nodes involved,
but the validation method used in the approach are not explained in details \cite{li2020blockchain}. Hence, no concrete comparative analysis could be performed with their approach. \cite{kang2020reliable} only reported the effects of adding reputation to the validation process in FL rather than detailing the impact of reputation to mitigate the adversities caused by malicious nodes making the experimental results unclear. Due to the lack of existing works on proof-of-stake (PoS) based blockchained FL and based on the studies mentioned above and other findings, we propose a PoS-based validator-voting scheme for blockchain-based FL, as an alternative approach to the existing works to dilute the adverse influence of malicious nodes in a blockchained FL system. 

% \textcolor{red}{Based on these studies, we propose a Proof-of-Stake based validator-voting scheme for blockchain-based FL which performs fairly or better than the existing state-of-the-art approaches to the best of our knowledge. [JH: We cannot claim that our proposed method is better than the baselines, without simulations; based on these studies" is not sufficient, and we should clarify what are the limitations of these works and how our proposed method resolves this issue]}\textcolor{blue}{What about we state the issues we found in the related papers and explain their contributions as clear as possible, and say we propose a PoS-based approach, as an alternative to the existing works?}\textcolor{red}{In what follows, it is unclear what are the system environments (or models) and what are the proposed operations. Please separate them correctly}\textcolor{blue}{Will come back to this}

\begin{table}[t]
  \centering
  \resizebox{\columnwidth}{!}{\begin{tabular}{c|l}
  \toprule
  Symbols   & \multicolumn{1}{c}{Meaning}   \\ \midrule
  ${d \in \mathcal{D}}$ & A device in the Device Set\\
  ${w \in \mathcal{W}}$ & A worker in the Worker Set\\
  ${v \in \mathcal{V}}$ & A validator in the Validator Set\\
  ${m \in \mathcal{M}}$ & A miner in the Miner Set\\
  $R_j$ & The $j$-th communication round\\
  $train_{d}$ & Private training set of device ${d}$\\
  $test_{d}$ & Private test set of device ${d}$\\
  $L_j^{d}$ & Locally updated model by device $d$ in round $R_j$\\
  $G_{j}$ & Global model constructed at the end of round $R_j$\\
  $tx_j^{w}$ & Worker-transaction containing $L_j^{w}$ initiated and signed by worker $w$ in $R_j$\\
  $A^{v}(L_j^{d})$ & Accuracy of $L_j^{d}$ evaluated by validator $v$ using test set $test_{v}$ in $R_j$\\
  $vt^{v}(L_j^{w})$ & Voting result on $L_j^{w}$ from validator $v$\\
  $tx_j^{v}(L_j^{w})$ & Validator-transaction containing $tx_j^{w}$ and $vt^{v}(L_j^{w})$ initiated by $v$ in $R_j$\\
  $vt^{m, \mathcal{V}}(L_j^{w})$ & Aggregated voting results from $\mathcal{V}$ on $L_j^{w}$, collected by $m$\\
  $r_j^{d}$ & Rewards to device $d$ in round $R_j$\\
  $block^m_j$ & The candidate block initiated by the miner $m$ in round $R_j$\\
  $block_j$ & The legitimate block produced by the \emph{winning-miner} in round $R_j$\\
  $KickR$ & Blacklist the $w$ who is identified as malicious for $KickR$ rounds\\
    \bottomrule
  \end{tabular}}
  \caption{List of Notations}
\end{table}
 
\section{Operations of VBFL}
%COMMENT\textcolor{cyan}{[JH: My suggestion is to quickly elaborate only mandatory descriptions for these generic operations at this point, while deferring the details to the supplementary materials. Our key point is only the Validation Mechanism of VBFL. We should present this section as soon as possible, and spend our space mostly for these sections; I also suggest to change the section titles as follows: Overall]}\textcolor{cyan}{[MB: this section looks too descriptive. Better use headings and bold text to ease the reading.]}\textcolor{blue}{Hang: Rewritten some parts and fixed typos after submission.}

% \textcolor{red}{I think passive voice makes more sense as this is decentralized and every node gets equal chance to operate the system}

VBFL is operated by a set of devices $\mathcal{D}=\left\{d_1, d_2, \cdots, d_n\right\}$, and, similar to Vanilla FL, carries out the learning processes through a sequence of communication rounds $\mathcal{R}=\langle R_1, R_2, R_j, \cdots \rangle$. In each round,
each device ${d \in \mathcal{D}}$ is assigned to one of the following roles: worker ${w \in \mathcal{W}}$, validator ${v \in \mathcal{V}}$, and miner ${m \in \mathcal{M}}$, where $|\mathcal{W}| + |\mathcal{V}| + |\mathcal{M}|=|\mathcal{D}|$, and  performs its function according to the  role assigned. 
The $id$ of a device is its public key, which is used to verify the signatures of transactions or blocks generated by the device. For consistency, we denote a learning device in Vanilla FL as $w$ as well, as a learning device in Vanilla FL plays the worker role in VBFL.

A local model update by worker $w$ in communication round $R_j$, denoted as $L_j^w$, is generated by performing local learning on the global model constructed in round $R_{j-1}$, denoted as $G_{j-1}$. In Vanilla FL, $L_j^w$ is directly used to compute $G_j$, whereas in VBFL, $L_j^w$ would go through a validation process performed by all ${v \in \mathcal{V}}$ before being used to construct $G_j$. In addition, worker $w$ computes its expected rewards $r_j^w$ for learning $L_j^w$ by following the reward mechanism of VBFL-PoS. Worker $w$ then encapsulates both $L_j^w$ and $r_j^w$ in a worker-transaction $tx_j^w$ signed by $w$'s private key, and sends $tx_j^w$ to a randomly associated validator. Each validator $v$ then obtains $tx_j^w$ from all of its associated workers, and broadcasts $tx_j^w$ to all the other validators. By doing so, each ${v \in \mathcal{V}}$ would have $tx_j^w$, for all ${w \in \mathcal{W}}$, so that it could vote on each $L_j^w$. Rewards $r_j^w$ are granted to $w$ if $L_j^w$ is voted \emph{Positive} in the aggregated voting results from $\mathcal{V}$ recorded in the legitimate block of $R_j$, denoted as $block_j$.
%The operations of a worker are illustrated in \textbf{Algorithm \ref{alg:worker_local_update}} in the supplementary material.

%COMMENT\textcolor{cyan}{[JH: A sentence cannot start with mathmatical notation. Please revise them throughout the paper]}\textcolor{blue}{[Hang: addressed]}

Validator $v$ receives a verification-reward, $r_j^{v\mbox{-}veri}$, by verifying the signature of one worker-transaction $tx_j^w$. If the signature of transaction $tx_j^w$ is verified, $v$ would extract $L_j^w$ from $tx_j^w$ and cast a vote on $L_j^w$, denoted as $vt^{v}(L_j^w)$, as being either \emph{Positive} or \emph{Negative}, according to the VBFL's validator voting mechanism. The voting mechanism requires each $v$ to perform one-epoch of local model update by using its own local training set $train_v$, which will be elaborated in the section of 
%\textbf{Validation Mechanism of VBFL}
\textbf{VBFL Validator Voting Mechanism}. Validator $v$ also receives a validation-reward, $r_j^{v\mbox{-}vali}$, by voting on one $L_j^w$. Then, $tx_j^w$, $vt^{v}(L_j^w)$, $r_j^{v\mbox{-}veri}$ and $r_j^{v\mbox{-}vali}$ would be encapsulated in a validator-transaction $tx_j^v(L_j^w)$ signed by $v$'s private key. Each $m$ then receives $tx_j^v(L_j^w)$ from its associated $v$, and broadcasts $tx_j^v(L_j^w)$ to all the other miners. By doing so, each ${m \in \mathcal{M}}$ would have $tx_j^v(L_j^w)$, for all ${v \in \mathcal{V}}$, and therefore each $m$ would have access to every voting result $vt^{v}(L_j^w)$ for all $L_j^w$, ${w \in \mathcal{W}}$, by every ${v \in \mathcal{V}}$. 
%The operations of a validator are illustrated in \textbf{Algorithm \ref{alg:validator_voting}} in the supplementary material.

Miner $m$ receives a verification-reward, $r_j^{m\mbox{-}veri}$, by verifying the signature of one $tx_j^v(L_j^w)$. If the signature is verified, $m$ would extract $vt^{v}(L_j^w)$ from $tx_j^v(L_j^w)$. For all the extracted $vt^{v}(L_j^w)$, $m$ would aggregate the voting results that vote on the same $L_j^w$ by each ${v \in \mathcal{V}}$ into $vt^{m, \mathcal{V}}(L_j^w)$. Then, all the aggregated voting results $\{vt^{m, \mathcal{V}}(L_j^w)\}$, for all ${w \in \mathcal{W}}$, are put inside of a privately constructed candidate block $block_j^m$. The candidate block would also contain all the expected rewards $r_j^w$, $r_j^{v\mbox{-}veri}$, $r_j^{v\mbox{-}vali}$ and $r_j^{m\mbox{-}veri}$. Miner $m$ then mines its own candidate block according to the VBFL-PoS consensus by hashing the entire block content and signing the hash by its private key. This is equivalent to the 0-difficulty PoW mining.

For example, consider two workers $\mathcal{W}=\left\{w_1, w_2\right\}$ and three validators $\mathcal{V}=\left\{v_1, v_2,  v_3\right\}$ in $R_j$. After $w_1$ generates $L_j^{w_1}$, $L_j^{w_1}$ would be obtained by each ${v \in \mathcal{V}}$ due to the broadcast among $\mathcal{V}$, and the same for $L_j^{w_2}$. Suppose all the validators have voted on both $L_j^{w_1}$ and $L_j^{w_2}$, and $vt^{v_1}(L_j^{w_1})$ = $vt^{v_3}(L_j^{w_1})$ = \emph{Positive}, $vt^{v_2}(L_j^{w_1})$ = \emph{Negative}. Suppose there are two miners $\mathcal{M}=\left\{m_1, m_2\right\}$. After $m_1$ and $m_2$ aggregate the voting results, both $vt^{m_1, \mathcal{V}}(L_j^{w_1})$ and $vt^{m_2, \mathcal{V}}(L_j^{w_1})$ would be equal to $\{2 \emph{Positive}, 1 \emph{Negative}\}$. Similarly, assume $vt^{v_1}(L_j^{w_2})$ = $vt^{v_2}(L_j^{w_2})$ = $vt^{v_3}(L_j^{w_2})$ = \emph{Negative}, which results in $vt^{m_1, \mathcal{V}}(L_j^{w_2})$ = $vt^{m_2, \mathcal{V}}(L_j^{w_2})$ = $\{0 \emph{Positive}, 3 \emph{Negative}\}$. Finally, $m_1$ puts $vt^{m_1, \mathcal{V}}(L_j^{w_1})$ and $vt^{m_1, \mathcal{V}}(L_j^{w_2})$ into its candidate block $block_j^{m_1}$, and $m_2$ puts $vt^{m_2, \mathcal{V}}(L_j^{w_1})$ and $vt^{m_2, \mathcal{V}}(L_j^{w_2})$ into $block_j^{m_2}$.

Upon $block_j^{m}$ is mined, miner $m$ propagates this mined block to all the other miners in the network. After $m$ mines $block_j^{m}$ and receives all the propagated blocks in the network, by tracking the stake information recorded on its blockchain, $m$ picks the block generated by the miner possessing the highest stake (i.e., accumulated rewards) among $\mathcal{M}$, as the legitimate block. Under VBFL-PoS, only this legitimate block can be used to extract the rewards record and the voting results with the corresponding model updates. Each $m$ would append this legitimate block to its own blockchain, and request its associated $w$ and $v$ to download this legitimate block to append to their blockchains as well. In practice, instead of receiving all the propagated blocks, miners have to set a time limit to accept the propagated blocks (denoted as \emph{propagated-block-wait-time}) in case some blocks have a long transmission delay. 
%A miner's operations are illustrated in \textbf{Algorithm \ref{alg:miner_procedure}} in the supplementary material.

After a $d$ being $m$, $v$ or $w$ appends a block, $d$ processes the appended block by performing the following two tasks: (1) compute $G_j$ by using those model updates with the number of \emph{Postive} votes no less than the number of \emph{Negative} votes, and (2) update each device's stake by accumulating the recorded and qualified rewards. For example, assume the legitimate block was mined by $m_2$ as it possesses more stake than $m_1$, then each $d$ would receive $vt^{m_2, \mathcal{V}}(L_j^{w_1})$ = $\left\{2 \emph{Positive}, 1 \emph{Negative}\right\}$ and $vt^{m_2, \mathcal{V}}(L_j^{w_2})$ = $\left\{0 \emph{Positive}, 3 \emph{Negative}\right\}$. In this case, only $L_j^{w_1}$ would be used for the construction of $G_j$. Each device updates the stake record for all $d \in \mathcal{D}$ being $w$, $v$ or $m$ in $R_j$ by accumulating qualified $r_j^w$, $r_j^{v\mbox{-}veri}$, $r_j^{v\mbox{-}vali}$ and $r_j^{m\mbox{-}veri}$ recorded in the appended block. The tracked stake records are used by miners to pick the legitimate block by comparing the stakeholding among the block miners. Also under VBFL-PoS, if an $L_j^w$ receives more \emph{Negative} votes than \emph{Positive} votes such as $L_j^{w_2}$ in the above example, the corresponding worker (i.e., $w_2$) would be flagged as being potentially malicious so that rewards (i.e., $r_j^{w_2}$) would not be granted to this worker. If $w$ is flagged as malicious for $KickR$ consecutive communication rounds, each $d$ should blacklist $w$ and reject all future communications from it. 
%The block processing of a device is illustrated in \textbf{Algorithm \ref{alg:block_procession}} in the supplementary material.

The above descriptions assume VBFL runs in its most ideal situation, which means all the transactions are well received and verified, all the nodes have stable connections with each other, and all the nodes have similar hardware capability. Therefore, each $m$ may end up with the same block content, excluding its signature. In a practical situation, the blocks mined by ${\mathcal{M}}$ may have different contents due to varying transmission delays incurred by transactions and diverse hardware capacities of devices. Moreover, as there may be purposeful attacks to the global model from malicious miners, such as forging fake voting results with fake validator signatures, it is critical to identify and only use the voting results in the legitimate block to complete global model updates, as the miner creating the legitimate block is deemed the most trustworthy miner among ${\mathcal{M}}$ under the VBFL-PoS consensus. The discussions of legitimate block and \emph{winning-miner} selection are in Section \textbf{VBFL-PoS Consensus Mechanism.} % A high-level overview of the described operations can be visualized in Fig. \ref{fig:f2_VBFL_overview}.

After a $d$ constructs $G_j$, $d$ concludes $R_j$ and enters $R_{j+1}$. The role of each $d$ is reassigned in $R_{j+1}$ based on the role-switching policy. 

% \begin{figure}[H]
%     \centering
%     \includegraphics[width=1.0\columnwidth]{f2_VBFL_overview.png}
%     \caption{Overview of the VBFL operations in one communication round.} 
%     \textcolor{cyan}{[JH: This figure is not well visualized; At least use only straight lines for visibility; Is this the simplest example? Otherwise, please simplfy this; This example should be clearly described in the main body, rather than simply referring to it]}
%     \textcolor{cyan}{[JH: Remove the Vanilla FL part that looks redundant and confusing; Given the limited space, write only the key phrases instead of writing full sentences; The figure focuses more on the standard decentralized operations, rather than our key new elements, i.e., the validation mechanism; At the top, add a trifold brace (or arrows) and specify one physical node is randomly appointed as a worker, validator, or miner.]} \textcolor{red}{[JH: This consumes too much vertical space without conveying much meaningful messages. Please make it more compact; Caption should be updated.]} \textcolor{blue}{Hang: Please see the updated figure after submission. New figure still takes some vertical space though, but the timeline of operations is well visualized in my opinion. Please advise}
%     \label{fig:f2_VBFL_overview}
% \end{figure}

\section{VBFL Validator Voting Mechanism}
%\section{Validation Mechanism of VBFL}
%COMMENT\textcolor{cyan}{revise the section title, as it implicitly downgrades the (minor or other major) contributions of this work}
%%COMMENT\textcolor{blue}{Hang: Is separating this to a new section ideal?}
%COMMENT\textcolor{red}{[JH: First describe that the standard validation in blockchain that only check the legitimacy of the transaction signature, followed by elaborating that we propose an enhanced validation that additionally identifies whether the transaction (i.e., model) is signed by a legitimate or malicious device. In this sense we call this as an enhanced validation mechanism.]} \textcolor{blue}{Hang: In my opinion, verifying/validating signature and model validation are two different things, as a malicious device can also sign valid signature with bogus model. Therefore, I think our model validation may not be treated as an enhancement from signature verification.}

%COMMENT\textcolor{cyan}{[JH: Let's not use the quotation marks throughout the papers.}\textcolor{blue}{Mostly addressed.}

% The VBFL Validator Voting Mechanism consists of three parts:
% \begin{enumerate}
%     \item $v$ votes on individually received $L_j^w$ recorded in $tx_j^w$.
%     \item $m$ aggregates the voting results for a specific $L_j^w$.
%     \item Blacklist the corresponding $d$ of $m$ who has been identified malicious for a networkwide pre-agreed number (denoted as $KickR$) of continous communication rounds.
% \end{enumerate}

% \subsection{Validator Voting}

The Validator Voting Mechanism aims to exclude those significantly distorted $L_j^{w}$ sent from malicious or compromised workers (e.g., using abnormal datasets for training or adding artificial noise to their models) from the construction of $G_j$ in $R_j$. Ideally, to validate $L_{j}^{w}$, a validator $v$ would like to compare the accuracy of $L_{j}^{w}$ against the accuracy of $G_{j-1}$, both evaluated with $test_{w}$, denoted as $A^{w}(L_j^{w})$ and $A^{w}(G_{j-1})$, respectively, and hypothesize that if $L_j^{w}$ is severely distorted, $A^{w}(L_j^{w})$ would reveal a steep accuracy drop compared to $A^{w}(G_{j-1})$. On the other hand, if $L_{j}^{w}$ is legitimately learned, $L_{j}^{w}$ would either yield higher accuracy or suffer from minor accuracy drop compared to $A^{w}(G_{j-1})$. However, since $v$ in VBFL would only obtain $L_{j}^{w}$ but neither $train_{w}$ nor $test_{w}$, and a directly uploaded value pair $\{A^{w}(L_j^{w})$, $A^{w}(G_{j-1})\}$ cannot be trusted by $v$, $v$ would be in no reliable way to obtain the exact values of both $A^{w}(L_j^{w})$ and $A^{w}(G_{j-1})$. A workaround for $v$ is to compare $A^{v}(L_j^{w})$ against $A^{v}(G_{j-1})$, and treat it as a proxy comparison to $A^{w}(L_j^{w})$ against $A^{w}(G_{j-1})$. The rational of such a proxy comparison is that if the data sets possessed by $w$ and $v$ follow a similar distribution, $A^{v}(L_j^{w})$ and $A^{w}(L_j^{w})$ should have little difference, and likewise for $A^{v}(G_{j-1})$ and $A^{w}(G_{j-1})$. As a result, $v$ may subjectively choose a threshold value, denoted as $vh_{j}^{v}$, as a tolerance of an accuracy-drop measurement between $A^{v}(G_{j-1})$ and $A^{v}(L_j^{w})$. If $A^{v}(G_{j-1})-A^{v}(L_j^{w}) > vh_{j}^{v}$, which means the accuracy drop exceeds $v$'s tolerance, $v$ would flag both $L_{j}^{w}$ and $w$ as potentially malicious, resulting in a \emph{Negative} vote for $L_j^{w}$. Otherwise, $v$ treats $w$ as legitimate and casts a \emph{Positive} vote. 

We implemented this validation method for ${\mathcal{V}}$ in VBFL to evaluate its effectiveness. In the experiments, we introduced three malicious $d$ along with 17 legitimate $d$ into the VBFL network. Although the role of each $d$ was reassigned in a new communication round, to focus on evaluating the effectiveness, we enforced the same role-combination to devices as 12 $w$'s, 5 $v$'s and 3 $m$'s in each round to guarantee that there was enough number of ${\mathcal{W}}$ and ${\mathcal{V}}$. When a malicious $d$ became a $w$, it would inject Gaussian noise of variance 1 to its legitimately learned local model parameters to simulate a distorted local model update sent to ${\mathcal{V}}$. We randomly sharded the \emph{MNIST} training set to all 20 ${d \in \mathcal{D}}$, and each $d$ used the entire \emph{MNIST} test set as its local test set.

\begin{figure}
    \centering
    \includegraphics[width=0.9\columnwidth]{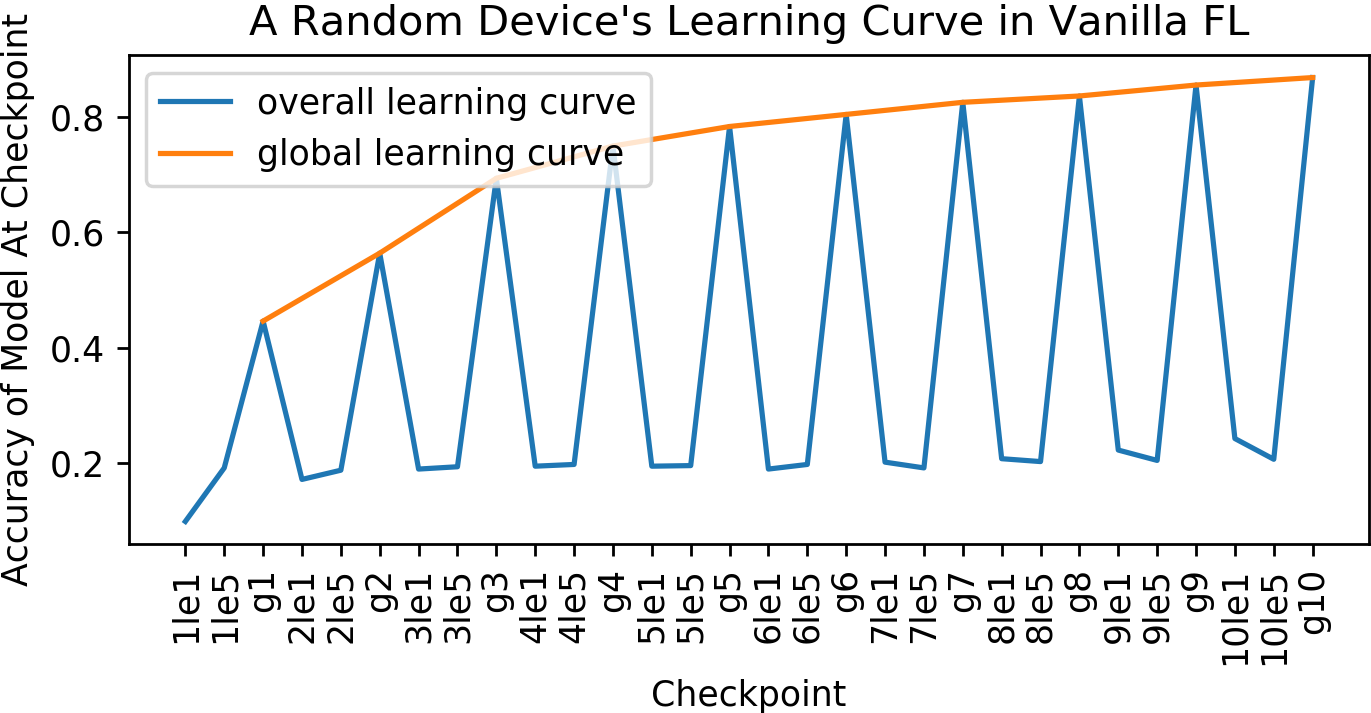}
    \caption{A Typical Device's Continuous Learning Curve in Vanilla FL Using \emph{FedAvg} \cite{mcmahan2017communication}}
    \label{fig:f3_device_overall_learning_curve_VFL}
\end{figure}

We randomly selected some fixed values for $vh_{j}^{v}$ (e.g., 0.01, 0.1, 0.2) and kept a log for both correctly and incorrectly identified malicious $w$ and their corresponding $A^{v}(G_{j-1})-A^{v}(L_j^{w})$ values. Unfortunately, this validation method could not effectively identify the malicious $w$, and the logged $A^{v}(G_{j-1})-A^{v}(L_j^{w})$ values exhibited an increasing trend for all ${w \in \mathcal{W}}$ and all ${v \in \mathcal{V}}$, showing virtually no distinctive relationship between $vh_{j}^{v}$ and $A^{v}(G_{j-1})-A^{v}(L_j^{w})$ for the legitimate and the malicious $w$'s. To investigate the reasons, we abided by the same data set sharding rule in the above VBFL experiments, and ran three Vanilla FL executions for 100 rounds with 20 legitimate $w$'s (i.e., learning devices) each performing 5 epochs of local training during each round and plotted their learning curves. Fig.~\ref{fig:f3_device_overall_learning_curve_VFL} displays the learning curve of one of the devices, say ${w}$, over the first 10 rounds. Below are the interpretations of Fig.~\ref{fig:f3_device_overall_learning_curve_VFL}:
\begin{itemize}
\item The peaks of accuracy values connected by the orange line represent $A^{w}(G_j)$, for $j = 1,2, ...,10$.
\item Let the locally updated model of ${w}$ after training $G_{j-1}$ for $n$ epochs in $R_j$ be denoted as $L_j^{w}(n)$. The accuracy values on the valleys correspond to the \emph{le} (i.e., local epoch) checkpoints. The accuracy at the point of \emph{jlen} on the x-axis represents $A^{w}(L_j^{w}(n))$.
\end{itemize}
Observed from Fig.~\ref{fig:f3_device_overall_learning_curve_VFL}, we concluded that our hypothesis was incorrect because a significant difference between $A^{w}(G_{j-1})$ and $A^{w}(L_{j}^{w})$ (i.e., $A^{w}(L_{j}^{w}(5))$, as $\mathcal{W}$ in this experiment perform 5 local epochs of training in each $R$) means that a legitimate $w$ also exhibits a sharp accuracy drop. The rest 90 rounds of $w$ and other legitimate learning curves show the same trend of accuracy drop. It is also observed that a sharp accuracy drop (i.e., from a peak to a valley) starts taking place when $w$ updates its local model from $G_{j-1}$ to $L_j^{w}(1)$. The range of this drop can be described as $A^{w}(G_{j-1}) - A^{w}({L_j^{w}(1)})$. 
%Our thoughts to the reasons behind such a phenomenon is expressed in the \textbf{Supplementary Materials}.

% \begin{itemize}
% \item The accuracy values on the peaks, which are connected by the orange line, corresponding to the \emph{g1}, \emph{g2}, ..., \emph{g10} checkpoint on the x-axis, represent the accuracy values evaluated by $G_{1}$, $G_{2}$, ..., $G_{10}$ on $test_{w_i}$(i.e., the whole \emph{MNIST} test set), denoted as $A^{w_i}(G_j)$, where $j = 1,2, ...,10$. For example, $A^{w_i}(G_{8})$ is 0.837. 
% \item The accuracy values on the valleys, which correspond to the $le$ (i.e., local epoch) checkpoint on the x-axis, represent the accuracy values evaluated by $w_i$'s locally updated model $L_j^{w_i}$ on $test_{w_i}$ at a certain epoch before replacing its $L_j^{w_i}$ with $G_{j}$. For example, the accuracy values corresponding to $le1$ and $le5$, before $g5$ and after $g4$ (i.e., 0.194 and 0.195), are evaluated by $w_i$'s $L_j^{w_i}$ after learning 1 epoch and 5 epochs from $G_{4}$. The ${w_i}$'s locally updated model after learning $n$ epochs in $R_j$ is denoted as $L_j^{w_i}(n)$ and the accuracy evaluated by $L_j^{w_i}(n)$ on $test_{w_i}$ is denoted as $A^{w_i}({L_j^{w_i}(n)})$. For example, $A^{w_i}({L_8^{w_i}(1)})$ is 0.207 and $A^{w_i}({L_7^{w_i}(5)})$ is 0.191.
% \end{itemize}

%COMMENT\textcolor{cyan}{[JH: The following paragraphs are a mixture of the proposed ideas and the rationale behind such designs. Please discuss one by one in a structural way, e.g., rationale followed by the proposed method, or the opposite way.]}\textcolor{blue}{Hang: Have been rewritten. Please take a look.}

To proceed with this observation, one possible solution is to get rid of $A^v(G_{j-1})$, and let validator $v$ first perform one-epoch of legitimate local learning and calculate $A^v(L_j^v(1))$ as a proxy evaluation of $A^w(L_j^w(1))$, with the assumption that $w$ is legitimately learning $L_j^w(1)$ as well. Validator $v$ would then calculate the validation accuracy difference $vad = A^v(L_j^v(1)) - A^v(L_j^w(n))$ and compare $vad$ with a newly defined validator-threshold value $vh_{j}^{v}$ to determine whether $L_j^w(n)$ is potentially distorted. By obtaining $L_j^v(1)$, $v$ would be able to approximate $L_j^w(1)$, bypassing the issue of accuracy drop as both $A^v(L_j^v(1))$ and $A^w(L_j^w(1))$ were produced after an accuracy drop from $G_{j-1}$. The hypothesis is that when a legitimate validator $v$ produces $A^v(L_j^v(1))$, the value of $vad = A^v(L_j^v(1)) - A^v(L_j^w(n))$ must be different between the $L_j^w(n)$ sent by a legitimate $w$ and that by a malicious $w$.

\begin{figure}[t]
    \centering
    \includegraphics[width=0.9\columnwidth]{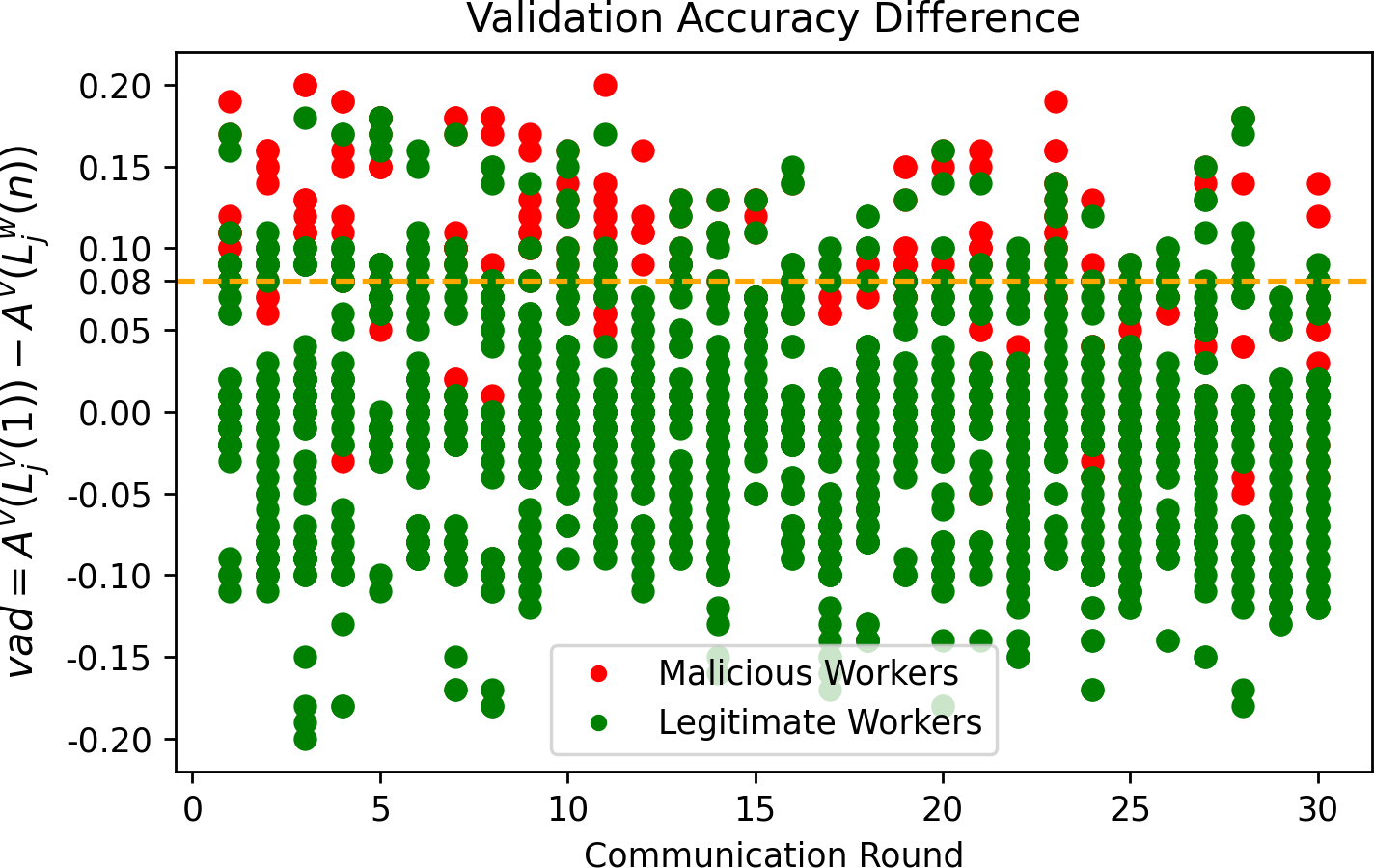}
    \caption{$vad$ values for both malicious $w$ and legitimate $w$.}
    \label{fig:f4_choice_of_vh}
    \vspace*{-0.1in}
\end{figure}

To find the pattern of $vad$ and its relationship with the new $vh_{j}^{v}$, we implemented this new validation method in VBFL and evaluated it using the exact same experimental setup with 3 malicious $d$ and logged the values of $vad$ for all ${w \in \mathcal{W}}$ and all ${v \in \mathcal{V}}$ across 30 communication rounds. To reduce complexity, when a malicious $d$ became a $v$, it did not execute any malicious action (e.g., flipping the voting result, etc.). Fig.~\ref{fig:f4_choice_of_vh} is a scatter plot that shows the values of $vad$ for every possible combinations of $v$ and $w$ for these 30 rounds. The red dots represent the values of $vad$ yielded by the malicious $w$'s and the green dots denoted those by the legitimate $w$'s. From Fig.~\ref{fig:f4_choice_of_vh} we observed that most red dots reside above the line $vad=0.08$. Therefore, if we let a $v$ considers the $w$ with $vad > 0.08$ as a malicious $w$ in each $R$, theoretically $v$ would be able to identify most of the distorted $L_j^w(n)$ and treat the corresponding $w$ as malicious. Here we can treat the 0.08 value as the newly defined validator-threshold $vh_{j}^{v}$, which defines the baseline of distinguishing maliciouos $w$ from legitimate $w$ when measuring $vad$. 

We evaluated the effectiveness of this validation method in Section \textbf{Experimental Results}. In summary, our proposed validator voting mechanism requires a validator $v$ in VBFL to perform one-epoch of local learning to get $A^v(L_j^v(1))$. After $v$ receives a $L_j^w(n)$, $v$ evaluates $L_j^w(n)$ on its test set $test_{v}$ to get $A^v(L_j^w(n))$, and calculates $vad = A^v(L_j^v(1))-A^v(L_j^w(n))$ to compare it with validator-threshold $vh_{j}^{v}$. If $vad > vh_{j}^{v}$, $v$ considers $L_j^w(n)$ to be potentially distorted and $w$ to be likely malicious, and generates a \emph{Negative} vote for $L_j^w(n)$. Otherwise, a \emph{Positive} vote for $L_j^w(n)$ is generated. The proposed validation mechanism on individual $L_j^{w}$ is presented in \textbf{Algorithm \ref{alg:validate_by_voting_mechanism}}.

\begin{algorithm}
	%\caption{ValidateByVotingMechanism (Used in \textbf{Algorithm \ref{alg:validator_voting}})}
	\caption{ValidateByVoting Mechanism}
	\label{alg:validate_by_voting_mechanism}
	For ${v \in \mathcal{V}}$\;
	Input: $L_j^w(n)$, $test_{v}$, $A^{v}({L_j^{v}(1))}$, $vh_{j}^{v}$\;
	 $r_{j}^{v\mbox{-}vali}$, $A^{v}(L_j^w(n))$ $\leftarrow$ $v$.\textit{\textbf{Evaluate($L_j^w(n)$, $test_{v}$)}}\;
	\eIf{$A^{v}({L_j^{v}(1))} - A^{v}(L_j^w(n)) > vh_{j}^{v}$}{
	 $vt^{v}(L_j^w(n))$ $\leftarrow$ Negative\;
	 }
	{
	 $vt^{v}(L_j^w(n))$ $\leftarrow$ Positive\;
	}
	\Return $r_{j}^{v\mbox{-}vali}$, $vt^{v}(L_j^w(n))$
\end{algorithm}

A limitation of finding $vh_{j}^{v}$ in our experiments is that $vh_{j}^{v}$ was determined by examining the emulation logs, while in practice, $\mathcal{V}$ would not have this ability in real time. We hypothesize that the value of $vh_{j}^{v}$ may be adaptively learned through examining all historical values of $A^{v}(L_{j'}^{v}(1)) - A^{v}(L_{j'}^{v}(n))$ when $v$ played the role of worker in $R_{j'}$, where $j'<j$, as the value of $A^{v}(L_{j'}^{v}(1)) - A^{v}(L_{j'}^{v}(n))$ would subjectively give $v$ a sense of approximating the $vad$ of a legitimate $w$. The study of $vh_{j}^{v}$ is deferred to future work.

\section{VBFL-PoS Consensus Mechanism}
%COMMENT\textcolor{cyan}{[JH: By contrast, this part is one key point that should be equally emphasized as the voting part, in terms of the depth, length, etc. This part also does not appear in the pseudo code, which should be clarified. Fig. 9 is only related to this, yet its impact on the final performance is still missing.]}\textcolor{blue}{[Hang: This part appears in the pseudo code in the supplementary materials. The section is also completely rewritten, please take a look.]}

\subsection{VBFL-PoS Reward Mechanism}
%COMMENT\textcolor{red}{[JH: This part is still unclear. Can we clarify at least the following key aspects using mathematical expressions?: (i) The mining winning condition under our PoS; (ii) the winning condition in the standard PoS; (iii) each entity's reward in the standard PoS; (iv) each entity's reward in our PoS]}\textcolor{blue}{[Hang: Added the mathematical expressions for rewarding mechanisms. The winning condition in my opinion is not expressable in clear math expression. It's a simple max() function in Python. I changed the PoS vs. PoS-without-worker-rewards in experimental results to PoS vs. PoW to show effectiveness of miner selection, as PoW does not take stake into account when selecting miner. The latency analysis between PoS and PoW is added in the supplementary material. Prof. Shen suggested we save the content in the supplementary material for a journal version.]}

%COMMENT\textcolor{red}{[JH: Similarly to Fig. 2, here let's add a figure that clarifies the rewards for a given PoS operational flow]}\textcolor{blue}{[Hang: Added the rewards flow in fig.2, please advise if it is good.]}

The VBFL-PoS Consensus Mechanism strives to protect the legitimately learned local model updates and ensure those updates are recorded on the blockchain and used for updating the global model. As miners are responsible for aggregating the voting results and recording them in a block, when a malicious device becomes a miner, it may try to disrupt the global model's calculation by putting bogus voting results and forged validator signatures in its mined block. As a result, avoiding choosing the block mined by a malicious $d$ is essential for robust blockchained FL.

To fulfill this purpose, inspired by the reward mechanism in BlockFL \cite{kim2019blockchained} and consolidated by the role switching policy, VBFL-PoS rewards devices according to the roles they play. Let $r$ denote a unit reward.

\begin{enumerate}
    \item Worker rewards: A worker $w$ in $R_j$ is rewarded proportionally to the number of data samples in $train_w$ and the number of local training epochs in $R_j$, denoted as $le_j^w$, \textbf{if} the number of \emph{Positive} votes for $L_j^w$, denoted as $N_{+}(L_j^w)$, is greater than or equal to the number of \emph{Negative} votes for $L_j^w$, denoted as $N_{-}(L_j^w)$, from $\mathcal{V}$ in $R_j$. The total rewards of worker $w$ in $R_j$ is calculated as:
    \begin{equation}
      r_j^w =
        \begin{cases}
         le_j^w * |train_w| * r & {if}~N_{+}(L_j^w) \geq N_{-}(L_j^w)\\
          0 & {if}~N_{+}(L_j^w) < N_{-}(L_j^w)
        \end{cases}       
    \end{equation}
    \item Validator rewards: A validator $v$ in $R_j$ is rewarded for verifying the signatures of the received worker-transactions $\{tx_j^w\}$ and voting for $\{L_j^w\}$ extracted from the signature-verified $\{tx_j^w\}$, namely, for generating $\{vt^v(L_j^w)\}$. The total rewards of $d$ being validator $v$ in $R_j$ is calculated as:
    \begin{equation}
    \begin{split}
      r_j^v & = |\{r_j^{v\mbox{-}veri}\}| + |\{r_j^{v\mbox{-}vali}\}| \\
      & = |\{tx_j^w\}| * r + |\{vt^v(L_j^w)\}| * r
    \end{split}
    \end{equation}
    Note that $|\{tx_j^w\}|$ is not necessarily equal to $|\{vt^v(L_j^w)\}|$. If the signature of a $tx_j^w$ is not verified, $v$ will not perform voting for the $L_j^w$ encapsulated in this non-verified $tx_j^w$.
    \item Miner rewards: A miner $m$ in $R_j$ is rewarded for verifying the signatures of received validator-transactions $\{tx_j^v(L_j^w)\}$, calculated as
    \begin{equation}
      r_j^m = |\{r_j^{m\mbox{-}veri}\}| = |\{tx_j^v(L_j^w)\}| * r
    \end{equation}
\end{enumerate}
\subsection{VBFL-PoS Miner Selection}
VBFL-PoS rewards worker $w$ with $le_j^w$ and $|train_w|$, if $L_j^w$ is positively voted, as a way to incentivize devices to contribute a large number of high-quality data samples and legitimately perform as many epochs as possible. As a result, the more rewards a device receives as a worker in $R_j$, the more positive contribution it has made to the training process in $R_j$. Workers are more heavily rewarded than validators and miners, as the verification and validation processes by validators and miners are regarded as obligation to keep VBFL-PoS running. Therefore, the accumulated stake of a device clearly shows the total contribution it has made to the entire learning process as the communication round progresses. When it comes to selecting the block for the global model update, VBFL-PoS dictates selecting the block mined by the miner possessing the highest stake among $\mathcal{M}$, as the miner with the highest stake has made the most contributions to the learning process among $\mathcal{M}$, and therefore it is regarded as the most trustworthy and the least probable to obstruct the learning process. The block selected by VBFL-PoS is called the legitimate block and denoted as $block_j$ in $R_j$. The miner producing $block_j$ is called the \emph{winning-miner} in $R_j$.

In the current implementation, there may be rounds where $\mathcal{M}$ consists of all malicious devices, or a malicious device is selected as the wining-miner because it possesses more stake than other legitimate miners in $\mathcal{M}$. The role-switching policy ensures miners to be re-selected in each new round, which reduces the chance where a malicious device is consecutively assigned to a miner role. Role-switching also prevents the “nondemocracy side effect" \cite{qu2019proof}, by which the device with the highest stake can be selected as the \emph{winning-miner} continuously. The prevention of such an effect can mitigate the risk that this device is compromised at a future point and start to attack the learning process. In cooperation with the blacklisting rule, the block mined by a blacklisted miner will get rejected by its associated devices, preventing VBFL-PoS from selecting that block as the legitimate block. We validated the effectiveness of miner selection under VBFL-PoS in the next section.

\begin{figure}[tbp]
    \centering
    \includegraphics[width=0.9\columnwidth]{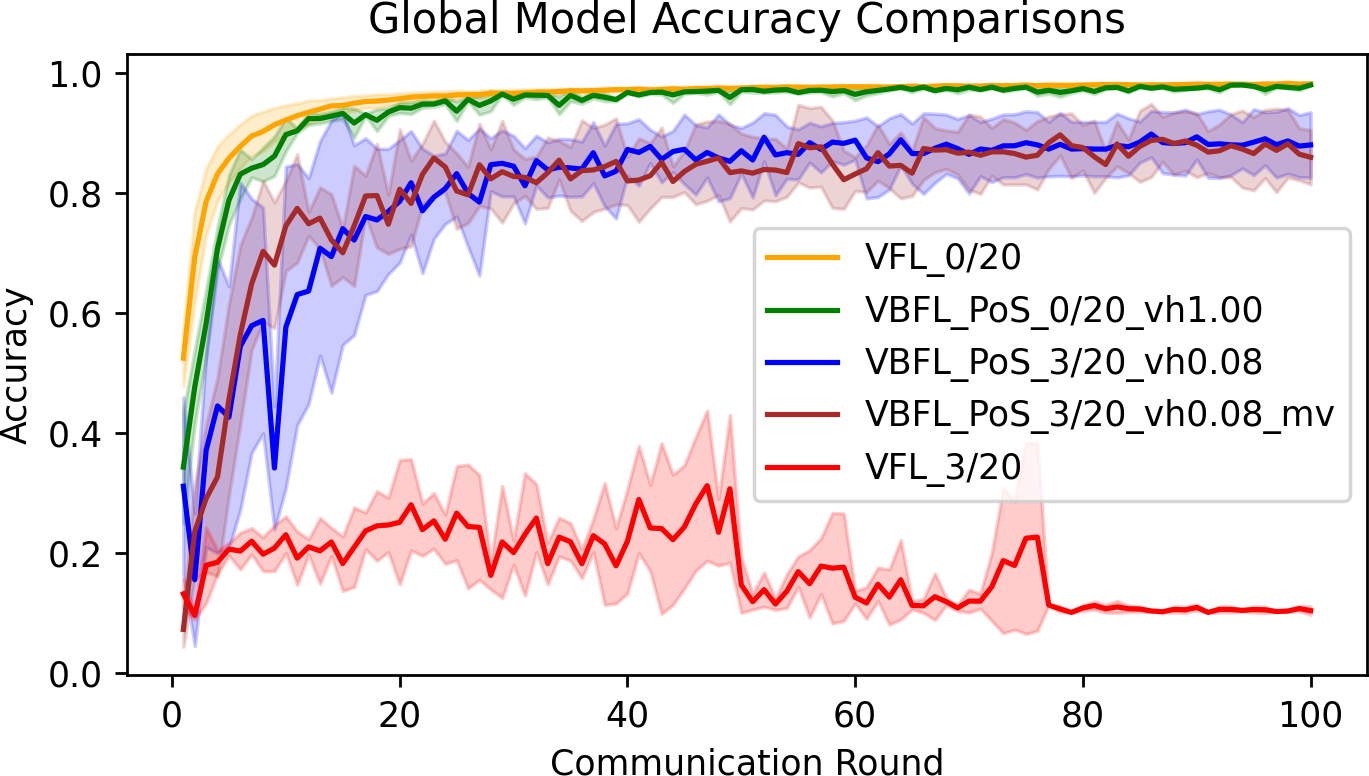}
    \caption{Effectiveness of Validation Mechanism}
    \label{fig:f5_effectivesness_validation_mechanisms}
\end{figure}

\begin{figure*}[t]
    \includegraphics[width=\textwidth]{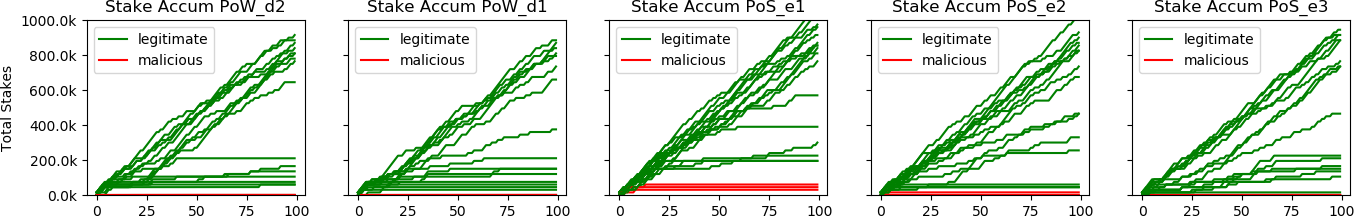}
    \caption{Stake Accumulation Curve}
    \label{fig:f9_PoS_stake_curve}
    \vspace*{-0.1in}
\end{figure*}

\section{Experimental Results}

%COMMENT\textcolor{cyan}{[JH: Please name the algorithms clearly, e.g., Vanilla FL, VBFL (mal. ratio: 3/20), VBFL (mal. ratio: 0/20), where mal. ratio is malicious device ratio; Remove On/Off, and they should be, e.g., with role-switching, without role-switching.]} \textcolor{cyan}{[MB: in all your plots  y-axis keep "accuracy" remove the remaining text.]} \textcolor{blue}{[Hang: both addressed.]}

All the experiments were conducted on a virtual machine with one NVIDIA V100 GPU, two Intel(R) Xeon(R) CPUs @ 2.30GHz and 13.34 gigabytes of RAM. All experiments involved 20 devices for both Vanilla FL and VBFL, and their training sets were assigned with a randomly sharded equally-sized portion of the entire \emph{MNIST} training set without overlap. $KickR$ was set to 6 and unit reward $r$ was set to 1 in all VBFL experiments. The verification of signatures is outside the scope of this work, so that all signatures were assumed to be verified. Each device in both Vanilla FL and VBFL adopted \emph{FedAvg} and the \emph{MNIST\_CNN}\cite{mcmahan2017communication} network structure, with 5 local training epochs per communication round, learning rate 0.01, and batch size 10. 

With the above basic configuration, we name the following experimental setups specifically:
\begin{enumerate}
    \item \emph{VFL\_0/20}: Vanilla FL with all 20 legitimate learning devices
    \item \emph{VFL\_3/20}: Vanilla FL with malicious ratio 3/20 (i.e., 3 out of 20 devices are malicious)
    \item \emph{VBFL\_PoS\_0/20\_vh1.00}: VBFL-PoS with all 20 legitimate devices and fixed validator-threshold 1.0
    \item \emph{VBFL\_PoS\_3/20\_vh0.08}: VBFL-PoS with malicious ratio 3/20 and fixed validator-threshold 0.08
    \item \emph{VBFL\_PoW\_3/20\_vh0.08}: VBFL-PoW with malicious ratio 3/20 and fixed validator-threshold 0.08
\end{enumerate}
Other specific experimental setups will be elaborated case by case.
\subsection{Effectiveness of VBFL's Validation Mechanism}
The effectiveness of the proposed validation mechanism can be visualized in Fig.~\ref{fig:f5_effectivesness_validation_mechanisms}. We ran 3 executions over 100 communication rounds for each of the 5 experiments labeled in the figure and recorded the global model accuracy at the end of each round produced by all the 20 devices. We manually assigned an arbitrary combination of 12 $w$'s, 5 $v$'s and 3 $m$'s for all three VBFL executions across all 100 rounds. For both Vanilla FL and VBFL, as each device used the entire \emph{MNIST} test set and the global model was consistent over devices at the end of each communication round, the evaluated accuracy at every round was the same from each device as shown in Fig.~\ref{fig:f5_effectivesness_validation_mechanisms}. The solid curves show the average accuracy values at each round of the 3 executions, and the upper bounds of the shaded areas depict the means plus the standard deviations of the 3 accuracy values at each round, and the lower bounds depict the means minus the standard deviations.

\begin{itemize}
     \item \textbf{FL without malicious devices:} The orange curve area show the accuracy of \emph{VFL\_0/20}, and the green curve area show that of \emph{VBFL\_PoS\_0/20\_vh1.00}. When all $v \in \mathcal{V}$ chose validator-threshold 1.0, they would vote \emph{Positive} for every available local model update, meaning every available local model update would be used to calculate the global model. As both experimental setups did not involve any malicious device, the green curve area is observed to be almost overlapped with the orange curve area, indicating the FL part of VBFL functioned correctly. Given the same number of total devices, although there were fewer learning devices (i.e., the workers) in VBFL compared to Vanilla FL (12 vs 20), its accuracy was not sacrificed because the role-switching policy ensures that every $train_w$ has a chance to be learned. 
    \item \textbf{Effectiveness of Model Validation:} The red curve area show the accuracy of \emph{VFL\_3/20}, and the blue curve area show that of \emph{VBFL\_PoS\_3/20\_vh0.08}. To focus solely on the effectiveness of model validation, there was no malicious behavior performed when malicious devices were selected as either validator or miner. Compared to \emph{VFL\_3/20}, the Validation Mechanism shown by the blue curve area effectively discards the distorted local model updates from at most 3 malicious devices and has the overall accuracy approaching that of the orange and the green curves. However, there is a roughly 10\% gap after convergence between the orange/green curves and the blue curve. This is attributed to the fact that up to 15\% of the \emph{MNIST} training data owned by the 3 malicious devices was never adequately learned. 
    \item \textbf{Effectiveness of Decentralized Voting:} In the setup of \emph{VBFL\_PoS\_3/20\_vh0.08\_mv}, when a malicious device became a validator, this device would always flip its voting result after it executed \textbf{Algorithm \ref{alg:validate_by_voting_mechanism}} for a local model update, and this is the only implementation difference compared to \emph{VBFL\_PoS\_3/20\_vh0.08}. The accuracy with malicious validators is shown by the brown curve area, which almost overlaps with the blue curve area, indicating that a few malicious or compromised validators in each round would not disrupt the validation process as a whole.
\end{itemize}

\subsection{VBFL-PoS vs. PoW Consensus in Miner Selection}

As the miner selection mechanism in PoW is indifferent to the stakeholding of miners, we can evaluate the effectiveness of legitimate miner selection in VBFL-PoS by comparing the maliciousness of the selected \emph{winning-miners} in PoW and VBFL-PoS. To implement PoW-based VBFL (termed VBFL-PoW), we added the PoW-mining operation for miners, which is doing nonce increments to meet the specified PoW-difficulty. Also, whenever a miner successfully finds the nonce of its candidate block, it would append this block to its blockchain, propagate this block to other miners and require its associated devices to download this block; when a miner receives a propagated block with a valid signature, the miner would stop mining its own block, append the received block to its blockchain, and require its associated devices to download this propagated block. For VBFL-PoS, we set the \emph{propagated-block-wait-time} to unlimited so that each miner was able to finish mining its own block and receive the propagated blocks from all the other miners, and decide the legitimate block immediately after it had received the last propagated block. 

\begin{figure}
    \centering
    \includegraphics[width=0.9\columnwidth]{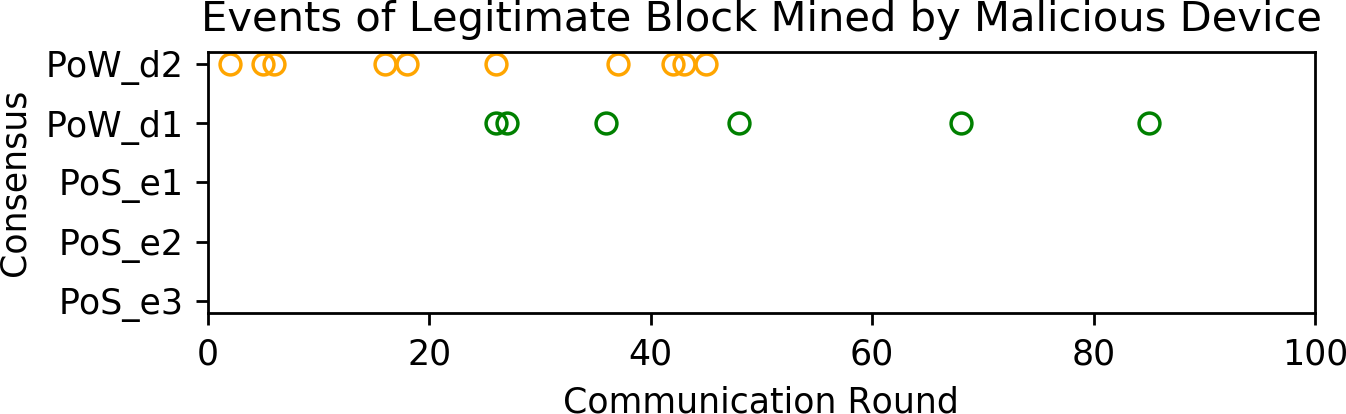}
    \caption{Effectiveness of VBFL-PoS Miner Selection}
    \label{fig:f8_malicious_miner}
    \vspace*{-0.1in}
\end{figure}

%Using the emulation logs of the five executions in the last subsection, 
In addition to the three VBFL-PoS executions in the last subsection, we ran one experiment of \emph{VBFL\_PoW\_3/20\_vh0.08} with PoW-difficulty 1 and another \emph{VBFL\_PoW\_3/20\_vh0.08} with PoW-difficulty 2. We plotted Fig.~\ref{fig:f8_malicious_miner} for each consensus setup, in which a circle indicates an event that a malicious device was selected as the \emph{winning-miner} during that communication round.
% The plot skipped plotting for the rounds in which a forking event happened or no valid block was mined. 
For the three VBFL-PoS experiments, no malicious device was selected as the \emph{winning-miner} in any round, while there were six rounds in PoW-difficulty 1 and ten rounds in PoW-difficulty 2, in which the legitimate block was mined by a malicious device. We also plotted the corresponding stake cumulation curves of all 20 devices in Fig.~\ref{fig:f9_PoS_stake_curve}. It is observed that the stake possessed by malicious devices increases very slowly and stopped increasing in an early stage in all five setups due to the validation mechanism. As a result, a malicious device was less likely chosen as the \emph{winning-miner} in VBFL-PoS, which explains the results of miner selection of the three VBFL-PoS experiments in Fig.~\ref{fig:f8_malicious_miner}. However, PoW did not take advantage of the stakeholding of the miners but the mining speeds, to decide the legitimate block, making it possible for a malicious device to be selected as the \emph{winning-miner} in any round, unless it is blacklisted by its associated devices.

\section{Conclusion and Future Work}
In this article, we have proposed VBFL, a blockchained FL framework that is robust against malicious distortions of local model updates. To mostly promote federation within only legitimate devices while excluding malicious devices, VBFL uses a decentralized validation mechanism to validate local model updates and a communication-efficient, FL-dedicated PoS consensus mechanism to select the block mined by the miner that has made the primary contribution to the learning process for extracting the positively voted local model updates to construct the global model. In our open-source prototype of VBFL, we have implemented various blockchain functionalities, such as peer registration, chain resyncing, the simulation of a forking event that are not effectively shown in this preliminary work. Utilizing such features, testing advanced adversarial attacks other than model distortions, and improving the blockchain mechanisms could be interesting for future research. In addition, noticing that PoS significantly reduces blockchain computing latency and energy consumption compared to PoW, it is worth studying to integrate VBFL with more communication and energy-efficient distributed learning frameworks than \emph{FedAvg} considered in this work. Lastly, the VBFL framework presented in this work operates in its emulation mode, and a more efficient blockchain data structure, chain-resyncing algorithm, and fault tolerance mechanisms have to be developed to test the performance of VBFL in practical distributed systems.

%\clearpage
\bibliographystyle{aaai}
\bibliography{bibliography.bib}

\vfill
\pagebreak

\end{document}